\PassOptionsToPackage{table,dvipsnames}{xcolor}
\documentclass[sigconf,nonacm]{acmart}
\AtBeginDocument{%
  }

\renewcommand\footnotetextcopyrightpermission[1]{}

\usepackage{threeparttable}
\usepackage{booktabs}
\usepackage{multirow}
\usepackage{array}
\usepackage{makecell}
\usepackage{bm}
\usepackage{float}
\usepackage{soul}
\usepackage{tcolorbox}
\usepackage{wrapfig}
\usepackage{bibunits}

\DeclareRobustCommand{\eg}{\textit{e.g.} }
\DeclareRobustCommand{\ie}{\textit{i.e.} }

\defaultbibliographystyle{ACM-Reference-Format}
\defaultbibliography{arxiv-EmoSEM-2025}

\begin{document}

\begin{bibunit}[ACM-Reference-Format]
\title{EmoSEM: Segment and Explain Emotion Stimuli in Visual Art}


\author{Jing Zhang}
\email{hfutzhangjing@gmail.com}
\orcid{0009-0005-1590-5886}
\affiliation{
  \institution{Hefei University of Technology}
  \city{Hefei}
  \country{China}
}

\author{Dan Guo}
\email{guodan@hfut.edu.cn}
\authornote{Corresponding author.}
\orcid{0000-0003-2594-254X}
\affiliation{
  \institution{Hefei University of Technology}
  \city{Hefei}
  \country{China}
}

\author{Zhangbin Li}
\email{lizhangbin.mail@gmail.com}
\orcid{0009-0001-2227-8826}
\affiliation{
  \institution{Hefei University of Technology}
  \city{Hefei}
  \country{China}
}

\author{Meng Wang}
\email{eric.mengwang@gmail.com}
\orcid{0000-0002-3094-7735}
\affiliation{
  \institution{Hefei University of Technology}
  \city{Hefei}
  \country{China}
}

\renewcommand{\shortauthors}{Zhang et al.}

\begin{abstract}
 This paper focuses on a key challenge in visual emotion understanding: given an art image, the model pinpoints pixel regions that trigger a specific human emotion, and generates linguistic explanations for it. Despite advances in general segmentation, pixel-level emotion understanding still faces a dual challenge: first, the subjectivity of emotion limits general segmentation models like SAM to adapt to emotion-oriented segmentation tasks; and second, the abstract nature of art expression makes it hard for captioning models to balance pixel-level semantics and emotion reasoning. To solve the above problems, this paper proposes the \textbf{Emo}tion stimuli \textbf{S}egmentation and \textbf{E}xplanation \textbf{M}odel (\textbf{EmoSEM}) model to endow the segmentation framework with emotion comprehension capability. First, to enable the model to perform segmentation under the guidance of emotional intent well, we introduce an emotional prompt with a learnable mask token as the conditional input for segmentation decoding. Then, we design an emotion projector to establish the association between emotion and visual features. Next, more importantly, to address emotion-visual stimuli alignment, we develop a lightweight prefix adapter, a module that fuses the learned emotional mask with the corresponding emotion into a unified representation compatible with the language model. Finally, we input the joint visual, mask, and emotional tokens into the language model and output the emotional explanations. It ensures that the generated interpretations remain semantically and emotionally coherent with the visual stimuli. Our method realizes end-to-end modeling from low-level pixel features to high-level emotion interpretation, delivering the first interpretable fine-grained framework for visual emotion analysis. Extensive experiments validate the effectiveness of our model. Code will be made publicly available.
\end{abstract}




\maketitle

\section{Introduction}
\label{sec:intro}

Affective computing is drawing more and more interest from the research community. Since artworks can evoke stronger emotions than natural images \cite{silvia2005emotional}, art emotion understanding is a prominent research topic \cite{cowen2020universal, iigaya2021aesthetic}. The research remains a major challenge due to the subjectivity of human emotion and the abstract nature of art. In this work, we explore a novel and specific emotion understanding problem: given an artwork image and an emotional prompt, the system detects pixel-level visual stimuli that evoke human emotional response and provides a linguistic explanation.

\begin{figure}
  \centering
     \includegraphics[width=0.98\linewidth]{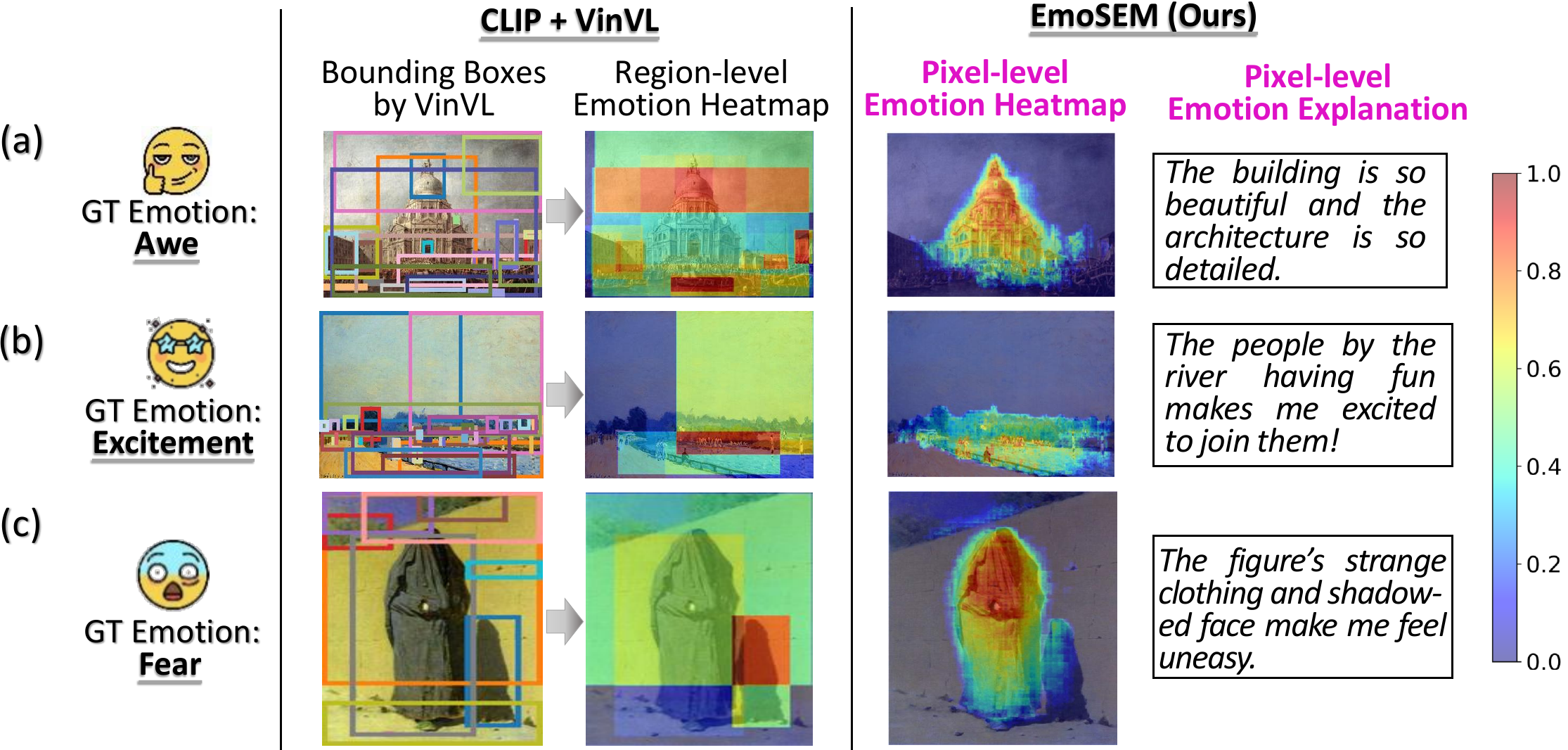}
    \caption{{Comparison of region- and pixel-level emotion stimuli. Region-level heatmaps represent discrete blocks, whereas pixel-level ones exhibit fine-grained, continuous variations, with each pixel assigned an emotion saliency value.}}
    \label{fig:bbox-vs-pixel}   
\end{figure}

In the field of visual emotion understanding, emotion recognition focuses on emotion classification \cite{chen2015learning,cen2024masanet, mohammad2018wikiart} or emotion distribution \cite{yang2022seeking,xu2021emotional}; emotional image captioning enhances descriptions with affective words (\eg, “lovely” or “alone”) \cite{mathews2016senticap,li2021image,wu2023sentimental}, and has expanded to tasks like sarcastic captioning \cite{ruan2024describe} and emotional visual dialog \cite{firdaus2020emosen,liu2024speak}. Furthermore, recent studies have begun exploring emotional explanation \cite{achlioptas2021artemis, zhang2024training}. The dataset \cite{achlioptas2021artemis} collects human emotion annotations and explanations, while \cite{zhang2024training} develops a small vision-language model for both emotion classification and explanation generation. However, these efforts primarily focus on the holistic relationship between an image and the viewer's emotional response. In reality, emotions are often triggered by specific visual elements, rather than by the image as a whole \cite{diffey2014tolstoy}. \cite{chen2024retrieving} takes a step forward by detecting visual stimuli that elicit emotional responses, yet focuses only on visual cues without reasoning. Stimulus detection and explanation are inherently complementary: what triggers the emotion and why. To this end, this work proposes a unified framework for emotion stimuli segmentation and explanation.

\begin{figure}  
  \includegraphics[width=0.98\linewidth]{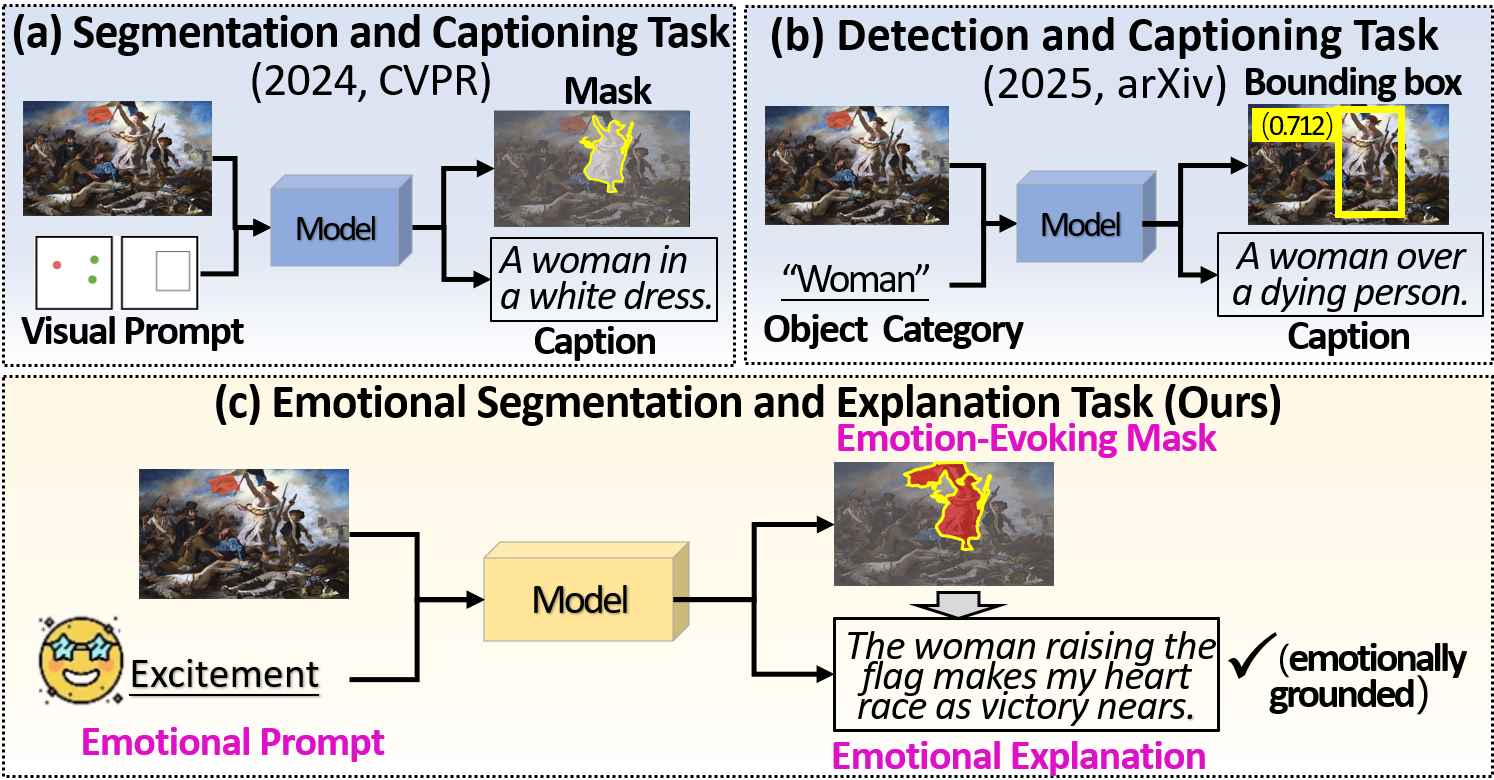}
  \caption{Our work focuses on the pixel-level emotional segmentation and explanation. Segmentation and captioning task (a) \cite{huang2024segment} and detection and captioning task (b) \cite{ren2024dino} represent similar architectures but target objective semantics. In contrast, we aim at emotion-oriented stimulus perception and reasoning, which is challenging due to the subjectivity of human emotions and the abstract nature of visual art.}
  \label{fig:task_comparison}
\end{figure}

Existing works \cite{zheng2017saliency,yang2018visual,yang2021solver,yang2021stimuli} employed the bounding boxes to localize emotion-relevant regions (\ie, emotional stimulus retrieval) for visual emotion recognition. As shown in Fig.~\ref{fig:bbox-vs-pixel}, region-level emotion stimuli rely on \textit{block-based activation}, which has two drawbacks. 1) It cannot accurately capture irregular shapes, such as the ``building'' in Fig.~\ref{fig:bbox-vs-pixel} (a) or the ``riverbank and people'' in Fig.~\ref{fig:bbox-vs-pixel} (b). 2) This way often encompass irrelevant visual content, resulting in inaccurate explanations. For example, in Fig.~\ref{fig:task_comparison} (b), the predicted bounding box unintentionally includes a ``dying person'' in the background, introducing conflicting with the ground-truth emotion ``excitement'' and distorting the interpretation.  In contrast, the pixel-level visual cues provide a more fine-grained representation of emotion by capturing \textit{continuous variations across pixels.} As shown in  Fig.~\ref{fig:task_comparison} (c), the pixel-level mask precisely highlights the ``woman'' and ``flag'' that express excitement, effectively removing unrelated background like the``dying person''. Therefore, we focus on pixel-level emotional analysis, which can identifies more precise emotion-triggering elements and enhances explanation consistency.

Our task faces two main challenges. Firstly, the general segmentation approaches focus on objective semantics and struggle to handle the emotional aspects. As shown in Fig.~\ref{fig:task_comparison} (a) and (b), segmentation and captioning task \cite{huang2024segment} depends on user-provided visual prompts (\eg, points, boxes), whereas detection and captioning task \cite{ren2024dino} operates bounding box localization based on object categories (\eg, ``woman''). In contrast, as shown in Fig.~\ref{fig:task_comparison} (c), our task is guided by subjective emotion prompts without any visual prompts and object categories, just with emotion prompts. 
Secondly and more importantly, beyond segmenting emotion stimuli, the model further possess the capability to interpret them. For this task, existing methods focus on objective region descriptions, outputting descriptions such as ``a woman in a white dress'' by SCA \cite{huang2024segment} and ``a woman over a dying person'' by DINO-X \cite{ren2024dino} irrelevant to the emotion aspects. Our task is fundamentally different. It requires to precisely discover the visual stimuli and describe the viewer’s emotional experience. As illustrated in Fig.~\ref{fig:task_comparison} (c), the explanation locates the emotion-evoking mask ``woman raising the flag'' and explains that from the viewer's emotional response ``makes my heart race as victory nears''. Ensuring alignment between the emotion-evoking mask and the explanation is the key challenge.

\begin{figure*}[!h]
        \centering
        \includegraphics[width=0.98\linewidth]{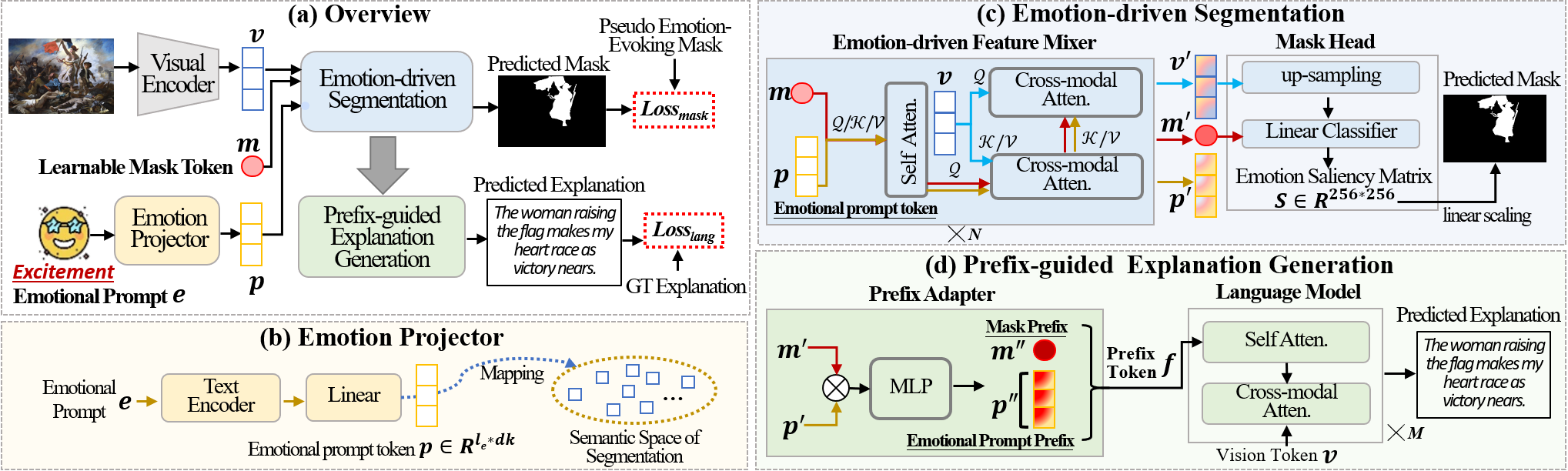}
        \caption{The network of our EmoSEM, comprising: (a) the overall architecture that is a visual segmentation–language generation pipeline; (b) an emotion projector that encodes the emotional prompt and embeds it into the semantic space of the mask decoder to facilitate emotion-aware segmentation; (c) the emotion-driven segmentation that predicts pixel-level emotion region by fusing visual and emotional cues with a learnable mask token modeling; (d) the prefix-guided explanation generation encoding masks and emotion embeddings as a prefix to guide the language model’s emotionally grounded interpretations.}
    \label{fig:framework}
\end{figure*}

To address the above issues, as illustrated in Fig.~\ref{fig:framework}, we design an Emotion stimuli Segmentation and Explanation Model (EmoSEM) for visual art understanding.
1) First, to fill the gap in segmentation models regarding emotional perception, we introduce an emotional prompt at the input stage of {segmentation}. We further design an emotion projector that embeds the emotional prompt into the {segmentation semantic space}, which, in conjunction with a learnable mask token and visual information, serves as the input. {This enables the model to integrate emotional cues with visual information for more accurate emotion-driven segmentation. 2) To enable meaningful emotional explanations for the masks, we devise a lightweight prefix adapter}. It transforms the emotional prompt features and the learnable mask token into a prefix token. This mapping converts the ``emotion-location” intent from the visual task into contextual signals that are interpretable by the language model. Finally, the prefix token is injected at the start of the language model’s decoding process, guiding it to generate explanations that are both semantically coherent and aligned with the mask’s emotional cues. 3) Besides, inspired by the observation that individuals may perceive different emotions from different regions of an image, we extend our model with a ``Single-Masks'' paradigm and a ``Multi-Masks'' paradigm (in Fig.~\ref{fig:two paradigms}). ``Multi-Masks'' enables concurrent processing of multiple emotion-specific masks, enhancing the flexibility and expressiveness of emotion segmentation.

In summary, our main contributions include:\begin{itemize}
\item We propose an Emotion stimuli Segmentation and Explanation Model (EmoSEM), providing a systematic solution for modeling emotion understanding from low-level visual perception to high-level emotional cognition. 
\item  {We introduce an emotional prompt along with the learnable mask token as input,  and design an emotion projector that effectively embeds the emotional prompt into the input space of the segmentation model, addressing the gap in emotion-guided visual segmentation.}
\item  {We further develop a simple yet effective {prefix adapter} that products a emotion \& mask-aware prefix token to bridge the segmentation model and the language model, facilitating that the generated explanations are more aligned with emotional stimuli.}
\item  Extensive experiments validate the effectiveness of our model, and showcase its broad applicability across single and multiple emotion-specific mask paradigms, enhancing the flexibility and expressiveness of emotion segmentation.
\end{itemize}

\section{Related Work} 
\label{sec:related_Work}

This section reviews two research areas relevant to our task, visual segmentation and visual emotion analysis. 

{\bf Visual Segmentation} {aims to group image pixels into semantic regions, evolving from CNN-based approaches \cite{simonyan2014very, long2015fully} to transformer-based models \cite{xie2021segformer, zheng2021rethinking}. Recent studies leverage vision-language models like CLIP \cite{radford2021learning} or ALIGN \cite{jia2021scaling} for open-world segmentation \cite{ghiasi2022scaling, liang2023open, shan2024open, lan2024proxyclip}. For examples, DINO-X \cite{ren2024dino} supports open-world object detection and captioning. In parallel, interactive segmentation has been explored with user inputs~\cite{xu2016deep,li2018interactive,zhao2024graco}, with SAM \cite{kirillov2023segment} as a pixel-level representative method.} However, these efforts focus on objective segmentation. Here, we shift to a subjective and emotion-driven setting, offering a novel perspective for the field.

{\bf Visual Emotion Analysis.} \textbf{``1) \textit{Visual emotion recognition}''} is a classical task, typically divided into emotion classification \cite{chen2015learning,cen2024masanet, mohammad2018wikiart} and emotion distribution learning \cite{yang2022seeking,xu2021emotional}. \textbf{``2) \textit{Emotional image captioning}''} aims to generate emotional captions by incorporating affective words (\eg, ``lovely'' or ``alone'') \cite{mathews2016senticap,li2021image,wu2023sentimental}. It has extended to sarcastic captioning \cite{ruan2024describe} and emotional visual dialog \cite{firdaus2020emosen,liu2024speak}. A few studies have explored the \textbf{``3) \textit{emotional visual explanation}''}. The ArtEmis dataset \cite{achlioptas2021artemis} captures emotions evoked by artworks along with their underlying reasons. \cite{zhang2024training} propose a small emotional vision-language model balancing performance and efficiency. \textbf{``4) \textit{Emotional stimulus retrieval}''} has been recently introduced, supported by the APOLO \cite{chen2024retrieving} with pixel-level emotional stimuli annotations, though focusing solely on visual cues without deeper reasoning. For emotion understanding, stimulus retrieval and explanation are complementary: the former identifies `\textit{what}'' trigger emotions, while the latter reveals ``\textit{how}'' perception is shaped. Building on this, we propose a unified framework that jointly segments and interprets pixel-level emotional stimuli, linking fine-grained visual elements to human emotions.

\section{Method}
\label{sec:method}
This section presents EmoSEM, a systematic solution for modeling fine-grained emotion understanding from low-level perception to high-level cognition. We first provide an overview of EmoSEM. Next, we detail the segmentation branch for pixel-level emotion localization, followed by the explanation branch for generating interpretable emotional explanations. Furthermore, we discuss the multi-task learning objectives and the different emotion paradigms.

\subsection{Overview for Our Task}
\label{sec:method-overview}
{The overview of our EmoSEM is shown in Fig.~\ref{fig:framework} (a). The framework comprises four components: a visual encoder, an emotion projector, an emotion-driven segmentation module, and a prefix-guided explanation generation module. Given an art image $\bm{I}$ and an emotional prompt $\bm{e}$, the visual encoder extracts vision token, while the emotion projector encodes the emotional prompt into a representation compatible with the segmentation process. Together with a learnable mask token $\bm{m}$, vision token and emotional prompt features are jointly fed into the segmentation branch to predict the pixel-level emotion-evoking mask $\widehat{\bm{\mathcal{M}}}$, grounding the visual elements that trigger the specified emotion. Finally, the explanation branch learns a prefix token $\bm{f}$ from the mask and emotion cues, which guides a language model to generate the emotionally aligned interpretation $\widehat{\bm{\mathcal{X}}}$. We elaborate on the details of each module below.}

\subsection{{Emotion-driven Segmentation}}
\label{sec:method-segmentation}

Given an art image $\bm{I}$, we employ a Vision Transformer (ViT) style \underline{\textbf{visual encoder}} $E_I$ to extract the vision token $\bm{v}=E_I(\bm{I})$, with size $64 \times 64$ and 256 dimensions. To adapt the segmentation model to our emotion-guided setting, we introduce an emotional prompt $\bm{e}$ , \ie, ``\textit{generate the mask for the emotion \_\_}'' as the conditional input. We further introduce an \underline{\textbf{emotion projector}} $P_{E}$ to map $\bm{e}$ into segmentation semantic space. Specifically, we utilize the word embedding layer of the language model \cite{radford2019language} as the text encoder, yielding a text representation of size $l_e\times d_w$, where ${l_e}=8$ is the length of $\bm{e}$ and $d_w=768$. To align the text representation with the input space of the segmentation model, we further transform it using a linear layer. This process yields an emotional prompt token, formulated as $\bm{p}=P_E(\bm{e}) \in \mathbb{R}^{l_e*d_k}$, where $d_k=256$.

After obtaining the vision token $\bm{v}$ and emotional prompt token $\bm{p}$, we construct \underline{a learnable mask token} $\bm{m}$, which, together with them, is fed into the \underline{\textbf{emotion-driven segmentation}} module to predict an emotion-evoking mask $\widehat{\bm{\mathcal{M}}}$. Built upon the mask decoder structure of SAM, this module integrates emotional prompts and learnable tokens through an emotion-driven feature mixer and a mask head. The \underline{emotion-driven feature mixer} is a bidirectional Transformer with $N$ stacked blocks \cite{vaswani2017attention}. In each block, the prompt token and the mask token first interact via self-attention (where the concatenation $[\bm{p},\bm{m}]$ is input) to establish emotional and spatial priors, and then engage in bidirectional cross-attention (abbreviated as ${\rm BiCrossAtt}$) with the vision token ($[\bm{p},\bm{m}]$-to-$\bm{v}$ and vice-versa) to achieve comprehensive cross-modal fusion:
\begin{equation}\label{eq: BiCrossAtt}
	{\rm Blcok}([\bm{p},\bm{m}],\bm{v})\!\! =\!{\rm BiCrossAtt}({\rm SelfAtt}([\bm{p},\bm{m}]),\bm{v})),
\end{equation}
where ${\rm BiCrossAtt}$/${\rm SelfAtt}$ is implemented by multi-head attention; feed-forward network and residual connection are omitted here.

Then, the mask head upsamples the updated vision token $\bm{v'}$, followed by an MLP to transform updated mask token $\bm{m'}$ into a dynamic linear classifier to predict an \ul{emotion saliency matrix} $\bm{S}\in \mathbb{R}^{256*256}$, where each entry indicates the pixel’s probability of evoking emotion. The saliency matrix is then scaled and thresholded following the procedure of SAM to derive the \ul{predicted mask} $\widehat{\bm{\mathcal{M}}}$:
\begin{align}\left\{\begin{aligned}\label{eq: Mask Prediction}
	&[\bm{p'},\bm{m'}],\bm{v'} ={\rm Feature Mixer}([\bm{p},\bm{m}],\bm{v}) \\
       &\widehat{\bm{\mathcal{M}}}={\rm Mask Head}(\bm{m'},\bm{v'})
\end{aligned},\right.\end{align}
where $\bm{p'},\bm{m'},\bm{v'}$ represent the updated prompt, mask and vision tokens, respectively, and $\widehat{\bm{\mathcal{M}}} \in \mathbb{R}^{h*w}$ denotes the predicted emotion-evoking mask. Here, $h$ and $w$ are the hight and width of image.

\subsection{{Prefix-guided Explanation Generation}}
\label{sec:method-explanation}

To effectively integrate emotional and mask information and provide richer contextual cues for the subsequent explanation process, we introduce a \underline{\textbf{prefix adapter}}. This adapter is implemented as a multi-layer perceptron (MLP) network. The updated mask token $\bm{m'}$ and emotional prompt token $\bm{p'}$ are concatenated and passed through the MLP network to generate a prefix token $\bm{f}$:
\begin{equation}\label{eq:prefix token}
\underset{\text{(emotional prompt prefix)}}{\smash{\bm{p''}}}\!\!\!,\ 
\underset{\text{(mask prefix)}}{\smash{\bm{m''}}} \!\!\!\!\!\!\! =\!{\rm PrefixAdapter}(\bm{p'},\bm{m'}),
\end{equation}
where \underline{prefix token} $\bm{f} =[\bm{p''},\bm{m''}] \in \mathbb{R}^{l_f*d_h}$, and $l_f$ is the length of prefix token; and $\bm{p''}$ and $\bm{m''}$ are emotional prompt prefix and mask prefix, respectively. These prefixes inject emotion- and region-specific cues from predicted masks into the language model, enabling it to capture both spatial semantics and emotional context. As will be demonstrated in experiments, these prefixes are critical for generating coherent and emotionally aligned explanations.

In our case, the GPT-2 \cite{radford2019language} is selected as the \underline{\textbf{language model}}. During training, it takes art image feature $\bm{v}$, the prefix token  $\bm{f}$, and ground-truth explanation $\bm{\mathcal{X}}$ as input \footnote{In the filed of image captioning, both the GT caption and the image are provided as input during training, whereas during inference, only the image is used as input.}, and generates the predicted emotional explanation $\widehat{\bm{\mathcal{X}}}$. However, {GPT2, mainly consists of text-based self-attention layers
(where the concatenation $[\bm{f}, \bm{\mathcal{X}}]$ serves as input in our work), which is not originally designed for multi-modal inputs. To improve this, we add a \textit{cross attention}  ($[\bm{f},\bm{\mathcal{X}}]$-to-$\bm{v}$) in each block.} The process is formulated as:
\begin{equation}\label{eq:LM}
\widehat{\bm{\mathcal{X}}}={\rm LanguageModel}(\bm{v},[\bm{f}, \bm{\mathcal{X}}]),
\end{equation}
where $,[\bm{f}, \bm{\mathcal{X}}]$ represents the use of the prefix $\bm{f}$ as the initial input to guide the GPT-2 in generating an explanation that focuses on the emotional aspects and the predicted mask.

\subsection{Training Objective}
\label{sec:method-Training Procedures}

For emotional segmentation, we employ Dice loss \cite{milletari2016v} to improve segmentation accuracy, and Focal loss \cite{lin2017focal} to address the foreground-background class imbalance. The training objective for emotional segmentation is as follows:
\begin{equation}\label{eq:Lmask}
L_{mask}=L_{\rm Dice}(\bm{\mathcal{{M}}},\widehat{\bm{\mathcal{M}}}) + L_{\rm Focal}(\bm{\mathcal{M}},\widehat{\bm{\mathcal{M}}}),
\end{equation}
where $\bm{\mathcal{M}}$ and $\widehat{\bm{\mathcal{M}}}$ represent the pseudo emotion-evoking mask and predicted emotion-evoking mask, respectively. For emotional explanation, we use a cross-entropy loss as the language loss to generate emotion explanations:
\begin{equation}\label{eq:Llang}
L_{lang}=L_{\rm XE}(\bm{\mathcal{X}},\bm{\widehat{\mathcal{X}}}),
\end{equation}
Our model is trained by minimizing the sum of all losses:
\begin{equation}\label{eq:Ltotal}
L_{total}=L_{mask} + L_{lang}.
\end{equation}

\begin{figure}
  \centering
     \includegraphics[width=0.98\linewidth]{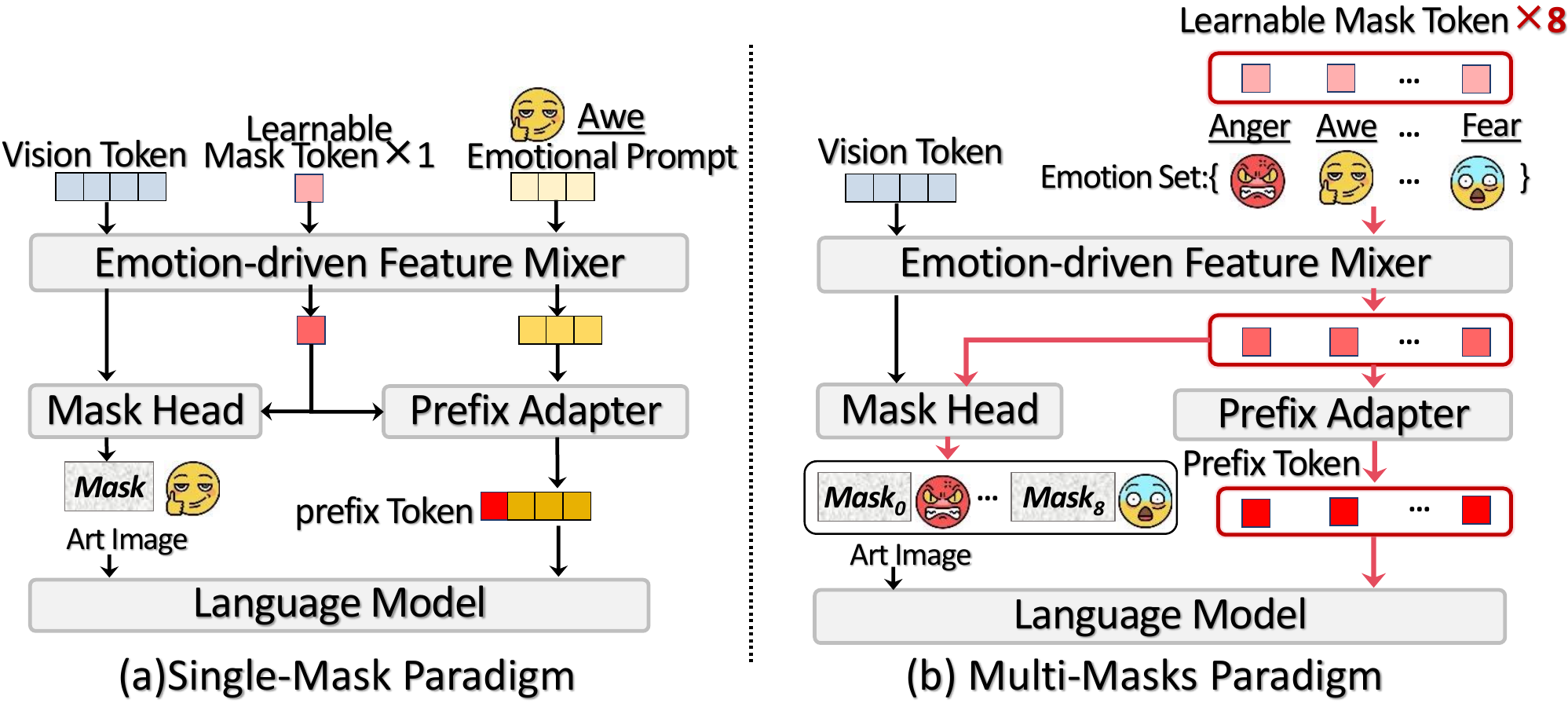}
    \caption{Our proposed framework supports two paradigms.
    (a) ``Single-Mask'' is our default paradigm with a single emotional mask output. (b)``Multi-Masks'' enables the simultaneous generation of multiple emotion masks for each image. It incorporate \underline{\textit{8 learnable mask tokens}}, each aligned with one of the eight fundamental emotions defined by Ekman \cite{ekman1992argument,bradley2007international}.}
    \label{fig:two paradigms}
\end{figure}

\begin{table*}[!h]
	\caption{Method comparison on the APOLO test set. 
    ``--” indicates that this result is not available.
    ``*” indicates results reproduced by us. The best and second best numbers in each column are marked with bold font and underlined, respectively. }
    \setlength{\tabcolsep}{2mm}
	\centering
        \begin{threeparttable}
    \begin{tabular}{m{1.8cm}|l|cc|cc|c|cccccc|c}
                    \toprule
                  \multirow{2}{*}{Type} & \multirow{2}{*}{Method} &  
                    \multicolumn{2}{c|}{Bounding Box}  & \multicolumn{2}{c|}{Segmentation}  & \multirow{2}{*}{EA$\uparrow$}  &  \multirow{2}{*}{B$_{1}$$\uparrow$} &   \multirow{2}{*}{B$_{2}$$\uparrow$}  & \multirow{2}{*}{B$_{3}$$\uparrow$} & \multirow{2}{*}{B$_{4}$$\uparrow$} & \multirow{2}{*}{M$\uparrow$} & \multirow{2}{*}{R$\uparrow$} & \multirow{2}{*}{CS$\uparrow$} \\
                     ~ & ~ & P@25$\uparrow$ & P@50$\uparrow$ & P@25$\uparrow$ & P@50$\uparrow$ & ~ & ~ & ~ & ~ & ~ & ~ & ~ & ~ \\
                    \midrule
                \multirow{6}{*}{\makecell[c]{Emotion Seg.}}  &  VilBERT \cite{lu2019vilbert} & 82.17 & 63.08 & 72.14 & 39.64 & -- & -- & -- & -- & -- & -- & -- & --  \\
                ~ & 12-in-1 \cite{lu202012} & 72.51 & 50.71 & 63.90 & 31.87 & -- & -- & -- & -- & -- & -- & -- & --  \\
               ~ &  CLIP+VinVL \cite{zhang2021vinvl}  & 81.97 & 63.00 & 71.29 &40.05 & -- & --  & -- & -- & -- & -- & --  & -- \\ 
                ~ & CLIP+SAM \cite{kirillov2023segment} *& 82.00 & 61.18 & 61.10 & 28.30 &  -- & --  & -- & -- & -- & - & -- & --  \\
                 ~ & TRIS \cite{liu2023referring} *  & 84.97 & 65.67 & 75.28 & 34.72 &  -- & --  & -- & -- & -- & - & -- & -- \\
                ~ & WESR \cite{chen2024retrieving} & 84.30 & 66.66 & 75.89 & \underline{44.97} & -- & -- & -- & -- & -- & -- & -- & -- \\
                \midrule
             \multirow{3}{*}{\makecell[c]{Emotion Seg.\\\& Exp.}}  &  DINO-X \cite{ren2024dino} * & {70.05} & {52.60} & {55.89} & {38.14} & {21.0} & {11.0} & {2.9}& {1.1} & {0.8} & {3.8} & {10.6} & {48.1}\\
    ~ &  SCA$_{base}$*  & \underline{86.00} & \underline{67.69} & \underline{77.93} & 44.62 & \underline{42.6} & \underline{16.2} & \underline{6.6} & \underline{3.0} & \underline{1.5} & \underline{6.7} & \underline{17.7} & \underline{55.1} \\
               ~ & \textbf{EmoSEM (ours)} & {\bf 86.29} & {\bf 68.65} & {\bf 78.06} & {\bf 45.45} & {\bf 56.6} & {\bf 21.4} & {\bf 8.8} & {\bf 4.0} & {\bf 2.0} & {\bf 8.4} & {\bf 18.2} & {\bf 56.8} \\ 
             \bottomrule
		\end{tabular}
   \end{threeparttable}
  \label{tab:method-comparison}
\end{table*}

\begin{table}[!h]
	\caption{{Effect of the proposed main modules.``$P_E$'' denotes Emotion Projector.} $\bm{m''}$ and $\bm{p''}$ represent mask prefix and emotional prompt prefix in Eq.~\ref{eq:prefix token}, respectively.}
    \setlength{\tabcolsep}{0.55mm}
	\centering
        \begin{threeparttable}
    \begin{tabular}{l|cc|cc|c|ccc|c}
                    \toprule
                    \multirow{2}{*}{Method} & \multicolumn{2}{c|}{Bbox.}  & \multicolumn{2}{c|}{Seg.} &  \multirow{2}{*}{EA} &  \multirow{2}{*}{B$_{4}$} & \multirow{2}{*}{M} & \multirow{2}{*}{R} & \multirow{2}{*}{CS} \\
                    ~ & P@25 & P@50 & P@25 & P@50 & ~ & ~ & ~ & ~ & ~ \\
                    \midrule
                    w.o $P_E$ & 85.66 & 67.39 & 77.56 & 43.04 & 37.1 & 1.7 & 8.1 & \underline{17.7} & 54.6 \\
                w.o $\bm{m''}$ \& $\bm{p''}$ & 86.16 & 67.63 & 77.71 & 43.19 & 12.4 & 1.2 & 8.0 & 16.8 & 54.6  \\
                w.o $\bm{p''}$ & 86.18 & 67.76 & {77.83} & \underline{44.68} & 52.6  & \underline{1.8} & 8.3 & \underline{17.7} & 55.0\\
                w.o $\bm{m''}$ & \underline{86.23} &\underline {67.95} & \underline{77.91} & {44.64} & \underline{52.8}  & 1.7 & \underline{8.4} & 17.6 & \underline{56.2} \\
                 \midrule
                 \textbf{full model}& {\bf 86.29} & {\bf 68.65} & {\bf 78.06} & {\bf 45.45} & \bf{56.6} & {\bf 2.0} & {\bf 8.4} & {\bf 18.2} & \textbf{56.8}\\ 
             \bottomrule  
		\end{tabular}
   \end{threeparttable}
  \label{tab:prefix-embedding}
\end{table}

\subsection{Different Emotion Paradigms}
\label{sec:method-Different Paradigms}

In this paper, we consider two emotional segmentation paradigms: 1) The \textbf{``Single-Mask''} paradigm is to generate a single emotion-evoking mask based on an emotional prompt $\bm{e}$ ``\textit{generate the mask for the emotion} \textunderscore \textunderscore'' with a fixed emotion. It has already been introduced in Fig.~\ref{fig:framework}. 2) Inspired by the observation that individuals may perceive distinct emotions from different regions of an image, we introduce a \textbf{``Multi-Masks''} paradigm to simulate the perception of multiple  distinct emotions simultaneously, enhancing segmentation flexibility. As illustrated in Fig.~\ref{fig:two paradigms} (b), \ul{\textit{we extend the learnable mask token to 8 tokens}} $[\bm{m_0},\!\cdots\!,\! \bm{m_7}]$, each corresponding to one emotion in: $\{$``amusement'', ``awe'', ``contentment'', ``excitement'', ``anger'', ``disgust'', ``fear'', ``sadness''$\}$ from \cite{ekman1992argument,bradley2007international}. Then, $[\bm{m_0},\!\cdots\!,\! \bm{m_7}]$ along with the vision token $\bm{v}$ are fed into the emotion-driven feature mixer, where mask tokens-based self-attention and two reverse cross-attentions (mask-to-vision and vice-versa) are performed. Subsequently, the updated mask token $\bm{m_i'}$ and vision token $\bm{v'}$ are then passed through the mask head to obtain the predicted masks: 
\begin{align}
\!\!\left
\{\begin{aligned}
\label{eq:Multi-Masks}
	&[\bm{m_0'},\!\cdots\!,\! \bm{m_7'}],\bm{v'}\!\! =\!\!{\rm Feature Mixer}([\bm{m_0},\cdots\!, \!\bm{m_7}],\!\bm{v}) \\
       &\widehat{\bm{\mathcal{M}_i}}={\rm Mask Head}(\bm{m_i'},\bm{v'}), \bm{i} \in \{0,\cdots,7\}
\end{aligned}\!,
\right.
\end{align}
where $\widehat{\bm{\mathcal{M}_i}}$ is the predicted emotion-evoking mask for the $\bm{i}$-th emotion category. Compared to the Single-Mask paradigm, the eight specific learnable mask tokens allow the model to more effectively segment distinct emotions. This hypothesis has been validated experimentally (see Tab.~\ref{tab:new-paradigm}).

For the \textbf{``Multi-Masks''} explanation generation, we map the updated mask token $\bm{m_i'}$ to the corresponding prefix token $\bm{f_i}$ through the prefix adapter, where $\bm{i} \in \{0,\cdots,7\}$. Then, we use the language model to generate emotional explanations:
\begin{align}\left\{\begin{aligned}\label{eq: Explanation Prediction}
	&\bm{f_i} ={\rm PrefixAdapter}(\bm{m_i'}), \bm{i} \in \{0,\cdots,7\} \\
       &\widehat{\bm{\mathcal{X}_i}}={\rm LanguageModel}(\bm{v},[\bm{f_i}, \bm{\mathcal{X}_i}])
\end{aligned},\right.\end{align}
where $\bm{\mathcal{X}_i}$ and $\widehat{\bm{\mathcal{X}_i}}$ denote the ground-truth explanation and predicted explanation for the $\bm{i}$-th emotion category, respectively. During training, the objective of Multi-Masks paradigm is implemented using Eq.~\ref{eq:Ltotal-emo}, which is the similar as that of Single-Mask paradigm. For each image training, the summation includes only the emotion categories annotated in its ground truth, excluding all unlabeled categories from the loss computation.
\begin{equation}\label{eq:Ltotal-emo}
L_{\text{total}} = 
\textstyle \sum_{emo = 0}^{7} L_{\text{mask}}^{emo} + 
\textstyle \sum_{emo = 0}^{7} L_{\text{lang}}^{emo}.
\end{equation}

\begin{table}[!h]
	\caption{Effect of different paradigms. ``Single-Mask'' is our default paradigm with a single emotional mask output, while ``Multi-Masks'' denotes the model enables the simultaneous generation of multiple emotional masks (refer to Fig.~\ref{fig:two paradigms}). The proposed framework outperforms under both paradigms.}
    \setlength{\tabcolsep}{0.45mm}
	\centering
        \begin{threeparttable}
    \begin{tabular}{l|cc|cc|c|ccc|c}
                    \toprule
                    \multirow{2}{*}{Paradigm}& \multicolumn{2}{c|}{Bbox.}  & \multicolumn{2}{c|}{Seg.}  & \multirow{2}{*}{EA}  & \multirow{2}{*}{B$_{4}$} & \multirow{2}{*}{M} & \multirow{2}{*}{R} & \multirow{2}{*}{CS} \\
                    ~  & P@25 & P@50 & P@25 & P@50 & ~ & ~ & ~ & ~ & ~ \\
                    \midrule
                
                Single-Mask &  {\bf 86.29} & \underline{68.65} & \underline{78.06} & \underline{45.45} & \bf{56.6}  & {\bf 2.0} & \underline{8.4} & {\bf 18.2} & {\bf 56.8} \\ 
                Multi-Masks  & \underline{86.09} & {\bf 68.98} & {\bf 78.13} & {\bf 46.68} & \underline{54.6} & \underline{1.9} & {\bf 8.7} & {\bf 18.2} & \underline{56.6}\\   
             \bottomrule                
		\end{tabular}
   \end{threeparttable}
  \label{tab:new-paradigm}
\end{table}

\section{Experiments}
\label{sec:Experiments}

\subsection{Experimental Setup}
\label{sec:Experiments-Setup}

\textbf{Datasets.} There is no pixel-level annotated dataset available \textbf{for training}. We use the training set of ArtEmis dataset~\cite{achlioptas2021artemis}. It is the largest available dataset for art emotion understanding, which contains 80,031 artworks with 454,684 affective responses and explanatory utterances (85$\%$/5$\%$/15$\%$ train/val/test). Since ArtEmis lacks mask annotations, following \cite{chen2024retrieving}, we generate pseudo pixel-level masks by detecting candidate regions with VinVL~\cite{zhang2021vinvl} and computing region–utterance similarities using CLIP~\cite{radford2021learning}. \textbf{For evaluation}, we adopt the APOLO dataset~\cite{chen2024retrieving}, which provides 6,781 manually annotated emotion-evoking masks for 4,718 artworks across 8 emotions (20$\%$/80$\%$ validation/test). Its emotion labels and explanatory utterances are derived from ArtEmis.

\begin{figure*}[h!]
  \centering
\includegraphics[width=0.98\linewidth]{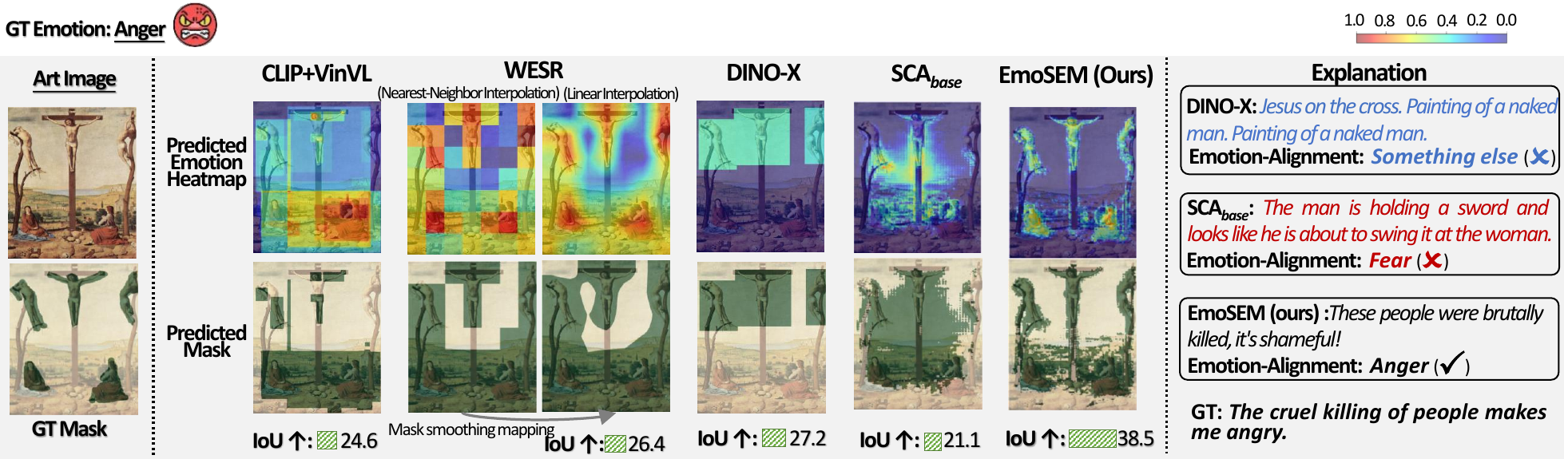}
     \caption{Example comparison on APOLO test set. We compare two segmentation models (CLIP+VinVL \cite{zhang2021vinvl}, WESR \cite{chen2024retrieving}), and  two segmentation\&explanation models
     (DINO-X \cite{ren2024dino}, the proposed baseline SCA$_{base}$ on SCA \cite{huang2024segment} for our task). IoU is the intersection over union of the predicted mask and the GT emotion-evoking mask, while the Emotion-Alignment metric \cite{achlioptas2021artemis} indicates how well the emotion category aligns with the emotion expressed in the explanations.  Our method performs best. }
    \label{fig:qualitative-results}
\end{figure*}

\begin{table*}
\centering
\caption{Computational efficiency comparison of EmoSEM across different backbone scales.}
 \setlength{\tabcolsep}{0.3mm}
\begin{tabular}{l|ccccc|cc|cc|c|ccc}
\toprule
\multirow{2}{*}{Method} &\multirow{2}{*}{\#Params} & \multirow{2}{*}{GPU} & \multirow{2}{*}{Training Time}   & \multirow{2}{*}{FLOPs$\downarrow$} &  \multirow{2}{*}{Inference Speed$\uparrow$} & \multicolumn{2}{c|}{Bbox.}  & \multicolumn{2}{c|}{Seg.} &  \multirow{2}{*}{EA$\uparrow$} &  \multirow{2}{*}{B$_{4}$$\uparrow$} & \multirow{2}{*}{M$\uparrow$} & \multirow{2}{*}{R$\uparrow$}  \\
~ & ~ & ~ & ~ & ~ & ~ & P@25$\uparrow$ & P@50$\uparrow$ & P@25$\uparrow$ & P@50$\uparrow$ & ~ & ~ & ~ & ~ \\
\midrule
SCA$_{base}$ & 237.67M & 1$\times$RTX 4090   & 47h, 10 epochs & 500.21G  & 5.91 (FPS) &  86.00 & 67.69 & 77.93 & 44.62 & 42.6 & 1.5 & 6.7 & 17.7 \\
EmoSEM-LLaVA   & 7B   & 4$\times$RTX 4090   & 30h, 3 epochs   & 5378.82G &  3.28 (FPS)  & 85.98 & 67.54 & 77.58 & 43.39 & 48.5  & 1.9  & \textbf{8.6}  & \textbf{18.2} \\
\textbf{EmoSEM-GPT2 (ours)}  & 276.72M  & 1$\times$RTX 4090  & 36h, 10 epochs & ~ {509.17G} &  {6.11 (FPS)}   & \textbf{86.29} & \textbf{68.65} & \textbf{78.06} & \textbf{45.45} & \textbf{56.6} & \textbf{2.0} & 8.4  & \textbf{18.2} \\
\bottomrule
\end{tabular}
\label{tab:emosem-llava}
\end{table*}

\begin{table}
\centering
\caption{{Comparing EmoSEM with GPT-4o and LLaVA-7B for pixel-level emotional explanation.}}
  \setlength{\tabcolsep}{1.6mm}
\begin{tabular}{lcccccccc}
\toprule
{Method} & {EA} & {B\textsubscript{1}} & {B\textsubscript{2}} & {B\textsubscript{3}} & {B\textsubscript{4}} & {M} & {R} & {CS} \\
\midrule
GPT-4o     & 51.0 & 12.5 & 3.0 & 1.0 & 0.0 & 5.0 & 11.5 & 52.1 \\
LLaVA-7B   & 44.0 & 13.5 & 4.8 & 2.2 & 1.2 & 5.2 & 12.4 & 53.7 \\
EmoSEM     & \textbf{57.0} & \textbf{24.3} & \textbf{11.5} & \textbf{6.1} & \textbf{3.5} & \textbf{9.5} & \textbf{19.3} & \textbf{55.4} \\
\bottomrule
\end{tabular}
\label{Comparing EmoSEM with GPT-4o and LLaVA-7B}
\end{table}

\textbf{Evaluation metrics.} {For \textbf{emotional segmentation},  we follow \cite{chen2024retrieving} to evaluate both bounding box \cite{peng2016emotions} and segmentation \cite{fan2018emotional} scenarios using precision at IoU thresholds (P@25, P@50). \textbf{For explanation}, BLEU (B), METEOR (M), and ROUGE(R) assess semantic relevance to the ground-truth captions, CLIPScore (CS) \cite{hessel2021clipscore} evaluates text–image similarity, and emotion alignment (EA) \cite{achlioptas2021artemis} measures consistency between the explanation’s emotion and the mask’s ground truth.}

\subsection{Main Evaluation}
{\bf How capable are existing segmentation models in detecting emotional stimuli?} We benchmark our method against existing artistic emotion segmentation approaches. Among the them, we reproduce CLIP+SAM and validate the recently proposed TRIS model \cite{liu2023referring}. As shown in Tab.~\ref{tab:method-comparison}, theses objective segmentation models \cite{zhang2021vinvl,lu2019vilbert,lu202012,liu2023referring, kirillov2023segment} do not transfer well to the emotion-oriented segmentation. In constrast, our method surpasses the SOTA method WESR \cite{chen2024retrieving}, \eg, in the segmentation scenario, achieving gains of 2.17$\%$ and 0.48$\%$ in P@25 and P@50, respectively.

{\bf Can our model be effective in both sentiment segmentation and interpretation?} 1) DINO-X \cite{ren2024dino} is a state-of-the-art pretrained open-world detector that outputs bounding boxes and region descriptions from textual prompts. We test its ability with emotion-related prompts. In Tab.~\ref{tab:method-comparison}, the low performance suggests that DINO-X is inadequate for emotion-oriented detection and description. 2) We further establish SCA$_{base}$ as a baseline, adapted from SCA~\cite{huang2024segment}, a SAM-based backbone, by integrating our emotional prompt and emotion projector and training it under identical settings. In Tab.~\ref{tab:method-comparison}, EmoSEM surpasses SCA$_{base}$ and other state-of-the-art methods across all segmentation metrics, while achieving a 1.7$\%$ gain in METEOR and a 14.0$\%$ EA improvement for explanations. These results confirm EmoSEM's strength in both segmentation accuracy and emotional explanation.

\begin{figure}
  \centering
    \includegraphics[width=0.98\linewidth]{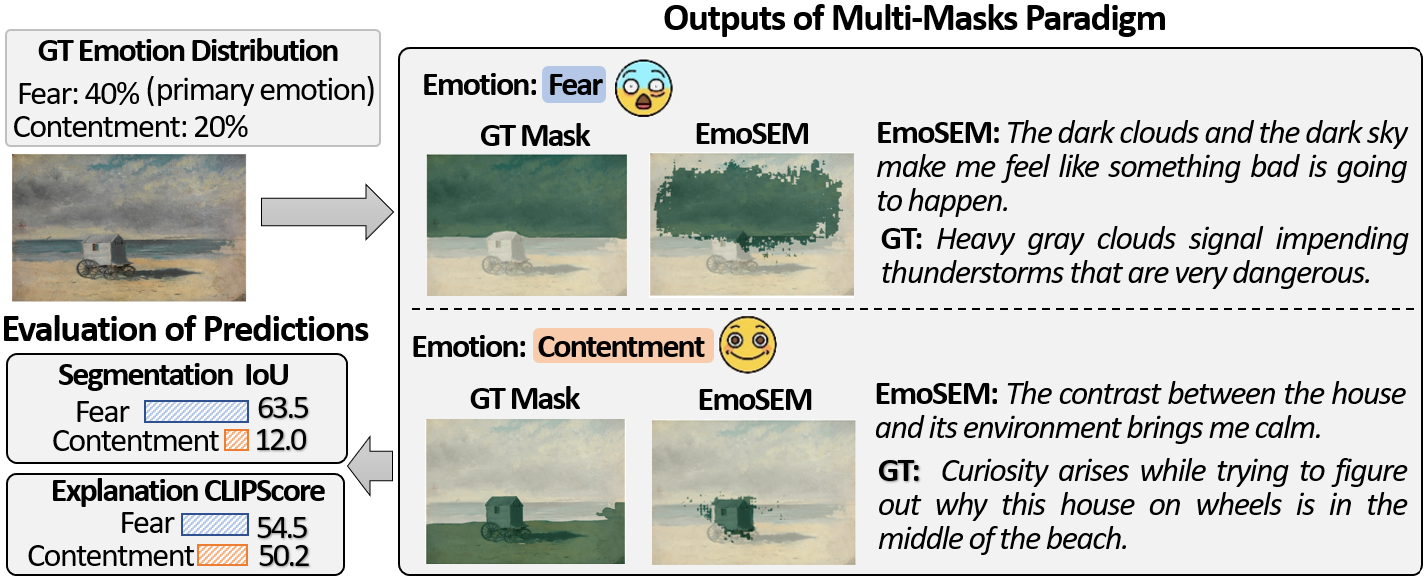}
    \caption{Multi-Masks paradigm visualization. This paradigm can accurately and simultaneously  detect emotion-specific visual stimuli corresponding to different individual emotions.}
    \label{fig:qualitative results of new paradigm}
\end{figure}

\subsection{Further Analysis}  
\textbf{Ablation study on main modules.} 1) In Tab.~\ref{tab:prefix-embedding}, replacing emotion projector with a raw word embedding module (\ie the ``w.o $P_E$" setup) degrades performance, confirming its effectiveness in mapping text prompts to the mask space. 2) Removing the mask (w.o $\bm{m''}$) or prompt prefix (w.o $\bm{p''}$) reduces EA by 3.8$\%$ and 4.0$\%$, respectively. 
When both are absent and replaced by the emotion word embedding, it performs worst in EA. In this case, the disconnect leads the language module to treat emotion cues as isolated text, impairing its ability to explain emotional stimuli at the pixel level. In contrast, the full model achieves the best EA of 56.6.

\noindent{\textbf{Paradigm comparison.} In Tab.~\ref{tab:new-paradigm}, both paradigms perform well, but differ in strengths. 1) Multi-Masks paradigm improves detection performance with P@50 gains of 0.33$\%$ and 1.23$\%$ in two scenarios. \textit{This is attributed to the eight dedicated different mask tokens} enabling more precise segmentation learning for each emotion (see Fig.~\ref{fig:two paradigms}(b)). 2) Single-Mask paradigm performs comparably in semantic quality but achieves 2.0$\%$ higher EA score. \textit{This might result from its explicit emotional prompt prefix $p''$ aiding emotion-text alignment}.}

\subsection{Discussion on MLLMs for Emotion  Stimuli}
Recent MLLMs like GPT-4o and LLaVA \cite{liu2023llava} show strong multi-modal comprehension, but none target emotion stimulus analysis in visual art. Here, we compare our EmoSEM against state-of-the-art MLLMs. 1) We compare EmoSEM with GPT-4o and LLaVA on 100 test samples due to GPT-4o's usage frequency limits (test samples to be released). Tab.~\ref{Comparing EmoSEM with GPT-4o and LLaVA-7B} shows that these MLLMs struggle to generate pixel-level emotion-grounded explanations, while our method shows notable advantages, \eg, +6.0\% EA over GPT-4o. 2) We further incorporate LLaVA as the language module in EmoSEM (forming EmoSEM-LLaVA). As shown in Tab.~\ref{tab:emosem-llava}, EmoSEM-LLaVA performs better than LLaVA but still lags in emotion segmentation and alignment compared to ours. More importantly, it has much larger size (7B, 25× ours at 276.72M) than ours. LLaVA pretrained objective corpora is still less suited for subjective tasks. 3) There is an interesting observation: existing MLLMs tend to force languages that fit the given emotion category based on linguistic style. They prefer to stack literary phrases over visual stimuli-grounded reasoning. In contrast, ours achieves better alignment among visual, emotion, and explanation. Please refer to Supplementary D.5.

\subsection{Qualitative Results}
We present a comprehensive visual analysis and summarize the following findings. 1) Fig.~\ref{fig:qualitative-results} shows our method achieves the best emotion segmentation, with masks closest to the ground truth. Regarding explanations, DINO-X \cite{ren2024dino} provides only objective descriptions without emotional content, while SCA$_{Base}$ with inaccurate segmentation results, leading to incorrect emotional explanations. In contrast, our method generates accurate emotional explanations.
2) In Fig.~\ref{fig:qualitative results of new paradigm}, the Multi-Masks paradigm can locate completely different visual stimuli for two specified emotions, \ie, “contentment” and “fear”. Moreover, the explanation for ``fear'' shows a higher CLIPScore (54.5) than that for ``contentment'' (50.2). This is consistent with the the model’s bias toward more dominant emotions.

\noindent{More implementation details and experimental results are provided in the supplementary material.}

\section{Conclusion}

This paper explores a novel and challenging task in the domain of artistic comprehension: emotional segmentation and explanation generation. To this end, we establish EmoSEM, a representative solution that enables a cross-modal leap from pixel-level perception to emotion-level reasoning. We conduct comprehensive experiments to validate the proposed modules and perform a thorough analysis comparing the emotion stimulus understanding of MLLMs. EmoSEM achieves state-of-the-art performance, demonstrating superior capabilities in both emotional localization and interpretation.

\putbib
\end{bibunit}

\newpage


\newcommand{\suptitle}{
  \twocolumn[{
  \centering
  \vspace{2em}
  {\Huge \bfseries EmoSEM: Segment and Explain Emotion Stimuli in Visual Art\par Supplementary Material\par}
  \vspace{2em}
  }]
}

\appendix
\suptitle

\begin{bibunit}[ACM-Reference-Format]

\section{Implementation Details}
The feature dimension $d_k$ of emotional prompt / learnable mask token is set to 256. For segmentation, the emotion-driven feature mixer includes $N=2$ blocks. For explanation generation, we choose GPT-2 \cite{sanh2019distilbert} as language model. It consists of $M=6$ transformer blocks and 12 attention heads. In the cross attention of language model, following \cite{zhang2024training,sanh2019distilbert}, we use CLIP ViT-B/16 \cite{radford2021learning} to extract visual features with a dimensionality of 768. The word embedding dimension $d_w$ is consistently set to 768. The explanations are truncated no longer than 25.
\textbf{During training}, the parameters of both the visual encoder and the feature mixer inherited from SAM \cite{kirillov2023segment} are kept frozen. The total batch size is 16 via 4-step gradient accumulation with a per-step batch size of 4. We adopt the AdamW optimizer \cite{gugger2018adamw} for all trainable parameters. The learning rate for the language model is 2e-4. For the Single-Mask paradigm, the emotion projector, mask head, and prefix adapter are collectively assigned a learning rate of 1e-4, whereas for the Multi-Masks paradigm, the learning rate is set to 8e-5. \textbf{During testing}, the segmentation visualizations use linear scaling to map the mask matrix to the image in this paper, and we use nucleus sampling \cite{holtzman2019curious} with a probability $0.9$ for explanation prediction. For detailed computational costs, please refer to Tab.~4 in the main paper. 

\section{Segmentation Models for Emotion Stimuli}
In Tab.~1 of the main paper, we benchmark our approach against existing methods for artistic emotion segmentation. These models link visual regions to emotion category, including 1) referring expressions models using emotion prompt as textual input (objective segmentation models VilBERT \cite{lu2019vilbert}, 12-in-1 \cite{lu202012}, TRIS \cite{liu2023referring} reproduced by us for this task), 2) CLIP-based emotion category-region similarity methods (CLIP+VinVL \cite{zhang2021vinvl}, CLIP+SAM \cite{kirillov2023segment} reproduced by us), and 3) a pixel-level emotion classification model trained with pseudo-labels (WESR \cite{chen2024retrieving}). As shown in Tab.~1 of the main paper, these methods \cite{zhang2021vinvl,lu2019vilbert,lu202012,liu2023referring, kirillov2023segment} perform suboptimally on the emotional segmentation. This suggests that objective segmentation models do not transfer well to the domain of emotion-oriented segmentation. Furthermore, our method surpasses WESR \cite{chen2024retrieving}. For example, in the segmentation scenario, EmoSEM achieves improvements of 2.17$\%$ and 0.48$\%$ in P@25 and P@50, respectively. 

\section{Proposed SCA$_{base}$ for Our Task}

\begin{figure}[!h]
        \centering
        \includegraphics[width=1\linewidth]{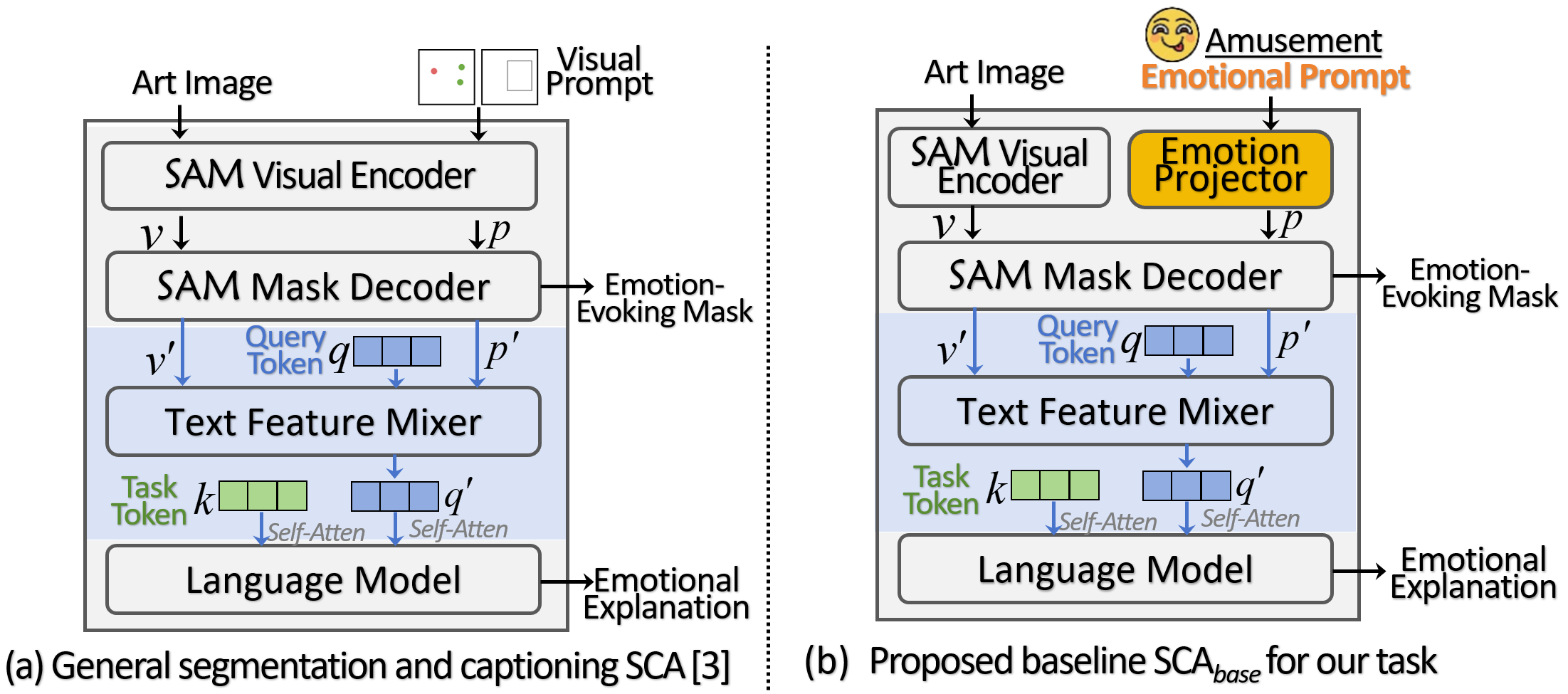}
        \caption{Overview comparison between (a) the general segmentation and captioning model SCA \cite{huang2024segment} and (b) the proposed baseline model SCA$_{base}$ for our task in this paper. SCA \cite{huang2024segment} is not inherently designed for emotion understanding; and we construct a baseline model, SCA$_{base}$ (b) with emotion projector, upon it.}
 \label{fig:overview-comparison}
\end{figure}

The task closely relevant to our study is segmentation and image captioning. SCA \cite{huang2024segment} is a representative method in this field. It primarily utilizes SAM \cite{kirillov2023segment} for image segmentation and a language model \cite{radford2019language} for general caption generation. As shown in Fig.~\ref{fig:overview-comparison} (a), SCA connects the SAM with the language model by introducing a learnable query token $q$, a learnable task token $k$, and a text feature mixer consisting of a 12-layer stacked bidirectional Transformer\cite{vaswani2017attention}. SCA \cite{huang2024segment} is not inherently designed for our task. To adapt it to our emotion-guided segmentation setting, in Fig.~\ref{fig:overview-comparison} (b), we introduce an emotional prompt along with the learnable token as the input of SAM decoder. And an emotion projector is devised for mapping the emotional prompt into the input feature space of SAM. We term this base model SCA$_{base}$.

\begin{table*}[!ht]
	\caption{Ablation study of different mask prefix strategies. 
    “Ambiguity-Aware” refers to concatenating multiple valid masks output by SAM to form the mask prefix for the language model, while “Ambiguity-Unaware” refers to using only one valid mask as the mask prefix. }
    \setlength{\tabcolsep}{1.8mm}
	\centering
        \begin{threeparttable}
    \begin{tabular}{l|cc|cc|c|cccccc|c}
                    \toprule
                    \multirow{2}{*}{Mask Prefix Strategy} & \multicolumn{2}{c|}{Bounding Box}  & \multicolumn{2}{c|}{Segmentation}  &  \multirow{2}{*}{EA}  &  \multirow{2}{*}{B$_{1}$} &   \multirow{2}{*}{B$_{2}$}  & \multirow{2}{*}{B$_{3}$} & \multirow{2}{*}{B$_{4}$} & \multirow{2}{*}{METEOR} & \multirow{2}{*}{ROUGE-L} & \multirow{2}{*}{CLIPScore} \\
                    ~ &  P@25 & P@50 & P@25 & P@50 & ~& ~ & ~ & ~ & ~ & ~ & ~ & ~ \\
                    \midrule
                 Ambiguity-Aware & \underline{86.23} & \underline{68.22} & \underline{77.95} & \underline{45.15} & \underline{55.6} & \bf{21.4} & \underline{8.6} & \underline{3.8} & \underline{1.8} & \bf{8.4} & \underline{17.8} & \underline{56.3} \\
                 \midrule
                 Ambiguity-Unaware (ours) & {\bf 86.29} & {\bf 68.65} & {\bf 78.06} & {\bf 45.45}& \textbf{56.6} & {\bf 21.4} & {\bf 8.8} & {\bf 4.0} & {\bf 2.0} & {\bf 8.4} & {\bf 18.2} & \textbf{56.8} \\ 
             \bottomrule  
		\end{tabular}
   \end{threeparttable}
  \label{tab:mask-prefix}
\end{table*}

\begin{table*}[!h]
\centering
\caption{Result of SimpleClick as segmentation architecture in EmoSEM.}
  \setlength{\tabcolsep}{1.85mm}
\label{tab:simpleclick}
\begin{tabular}{l|cc|cc|c|cccccc|c}
                    \toprule
                    \multirow{2}{*}{Segmentation Architecture} & \multicolumn{2}{c|}{Bounding Box}  & \multicolumn{2}{c|}{Segmentation}  &  \multirow{2}{*}{EA}  &  \multirow{2}{*}{B$_{1}$} &   \multirow{2}{*}{B$_{2}$}  & \multirow{2}{*}{B$_{3}$} & \multirow{2}{*}{B$_{4}$} & \multirow{2}{*}{METEOR} & \multirow{2}{*}{ROUGE-L} & \multirow{2}{*}{CLIPScore} \\
                    ~ &  P@25 & P@50 & P@25 & P@50 & ~& ~ & ~ & ~ & ~ & ~ & ~ & ~ \\
                    \midrule
SimpleClick & 85.98 & 67.54 & 77.58 & 43.39 & 51.0 & 19.9 & 7.9 & 3.6 & 1.7 & 7.9 & 17.5 & 56.0 \\
SAM (Ours) & \textbf{86.29} & \textbf{68.65} & \textbf{78.06} & \textbf{45.45} & \textbf{56.6} & \textbf{21.4} & \textbf{8.8} & \textbf{4.0} & \textbf{2.0} & \textbf{8.4} & \textbf{18.2} & \textbf{56.8} \\
\bottomrule 
\end{tabular}
\label{tab:SimpleClick}
\end{table*}

Our EmoSEM is fundamentally different from the baseline. SCA$_{base}$ primarily relies on the learnable query token $q$, which passes through a deep text feature mixer before being input into the language model for explanation generation (\emph{i.e.}, merely query context modeling for explanation generation). In contrast, EmoSEM adopts a more \textit{explicit} visual stimuli localization design. As illustrated in the Fig.~3 of the main paper, we first introduce a lightweight prefix adapter that enhances the mask token $m'$ and emotion prompt representation $p'$ into a prefix token aligned with the input space of the language model. Then the learned prefix token together with the art image is fed into the language model for explanation generation (\emph{i.e.}, explicitly utilizing visual, mask, and emotional tokens for explanation generation).

\section{More Experimental Results and Analysis}

\subsection{Fine-Tuning Design}

\begin{figure}[!h]
  \centering
\includegraphics[width=0.98\linewidth]{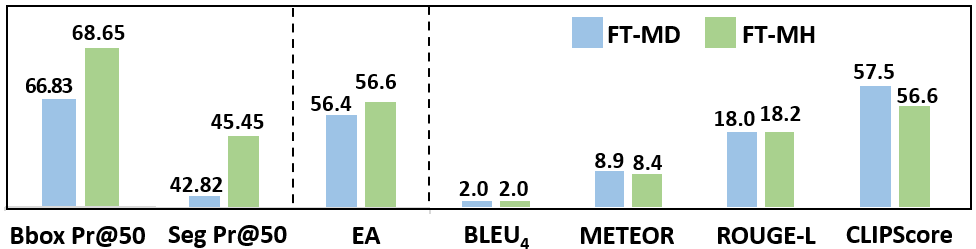}
    \caption{Fine-tuning strategies comparison. ``FT-MD'' means the \textit{entire} mask decoder is fine-tuned, and ``FT-MH'' signifies that \textit{only} the mask head is fine-tuned.}
      \label{fig:FTMaskDecoder}
\end{figure}

Fig.~\ref{fig:FTMaskDecoder} illustrates the comparative performance of two fine-tuning strategies (refer to the emotion-driven segmentation module in the Fig.~3 of main paper). The results demonstrate that fine-tuning the \textit{entire} Mask Decoder (FT-MD) does not yield significant improvements in semantic metrics. Moreover, this approach leads to a substantial decline in segmentation accuracy, exemplified by a 2.63$\%$ decrease in segmentation precision at IoU 50 (P@50). In contrast, fine-tuning \emph{only the mask head} (FT-MH) preserves segmentation performance while maintaining semantic understanding. Based on these observations, we adopt a strategy of fine-tuning \textit{only} the mask head (FT-MH) in all experiments.

\subsection{Different Mask Prefix Strategies}
The SAM architecture can generate multiple candidate masks for ambiguous inputs, referred to as the Ambiguity-Aware strategy \cite{kirillov2023segment}. Inspired by this, we discuss the strategy of single or multiple candidate masks in our model. In our default setting, we adopt an Ambiguity-UnAware strategy, which uses only a single default mask for simplicity and efficiency. We further investigate the Ambiguity-Aware strategy by incorporating multiple masks as mask prefixes. As shown in Tab.~\ref{tab:mask-prefix}, this design leads to performance degradation in most metrics, such as a 1.0$\%$ decrease in Emotional Alignment (EA) and a 0.4$\%$ drop in ROUGE-L score. This result suggests that introducing multiple ambiguous masks may increase noise and hinder the model’s ability to generate accurate and focused emotional explanations. Therefore, we adopt the Ambiguity-UnAware strategy in all experiments.

\subsection{Segmentation Backbone}
For the segmentation architecture, here we have tested another SOTA architecture, SimpleClick \cite{Liu_2023_ICCV}.  As shown in Tab.~\ref{tab:simpleclick}, EmoSEM built upon SAM architecture consistently outperforms its counterpart using SimpleClick across all evaluation metrics. This difference can be attributed to the architectural design. SimpleClick primarily relies on self-attention for feature interaction, which is effective for instance segmentation but less capable of leveraging cross-modal correlations required for emotion alignment. In contrast, SAM adopts a bidirectional cross-attention mechanism that facilitates stronger correspondence between emotional cues and visual features. Considering its superior performance and well-established robustness, we select the SAM architecture as the segmentation backbone in all experiments.

\subsection{Computational Efficiency Analysis}
Tab.~4 in main paper presents the computational cost comparison between the baseline and our EmoSEM model. Below, we provide a deep analysis to further explain the observed differences. 1) Our method exhibits slightly higher computational complexity in terms of FLOPs (509.17G vs. 500.21G), primarily due to the additional inclusion of \underline{a frozen CLIP encoder} \cite{radford2021learning} for visual feature extraction. Since this frozen module does not participate in gradient backpropagation, it imposes minimal overhead during training. 2) Although SCA has marginally lower FLOPs, its architecture includes a 12-layer bidirectional Transformer between the segmentation module and the language model as a \underline{text feature mixer}. This introduces intermediate computation steps such as layer normalization, and residual connections, along with frequent memory access during key-value retrieval and intermediate feature caching. \textit{These operations significantly increase training time and memory usage, resulting in longer actual training and inference times compared to our model, even though these are not reflected in FLOPs.} \textbf{In contrast, our \underline{lightweight prefix adapter} design, allows better hardware utilization and overall faster execution.} In summary, our method achieves faster training and lower inference latency, demonstrating the  efficiency of our model.

\subsection{More Visualization Analysis}

\begin{figure}
  \centering  \includegraphics[width=1\linewidth]{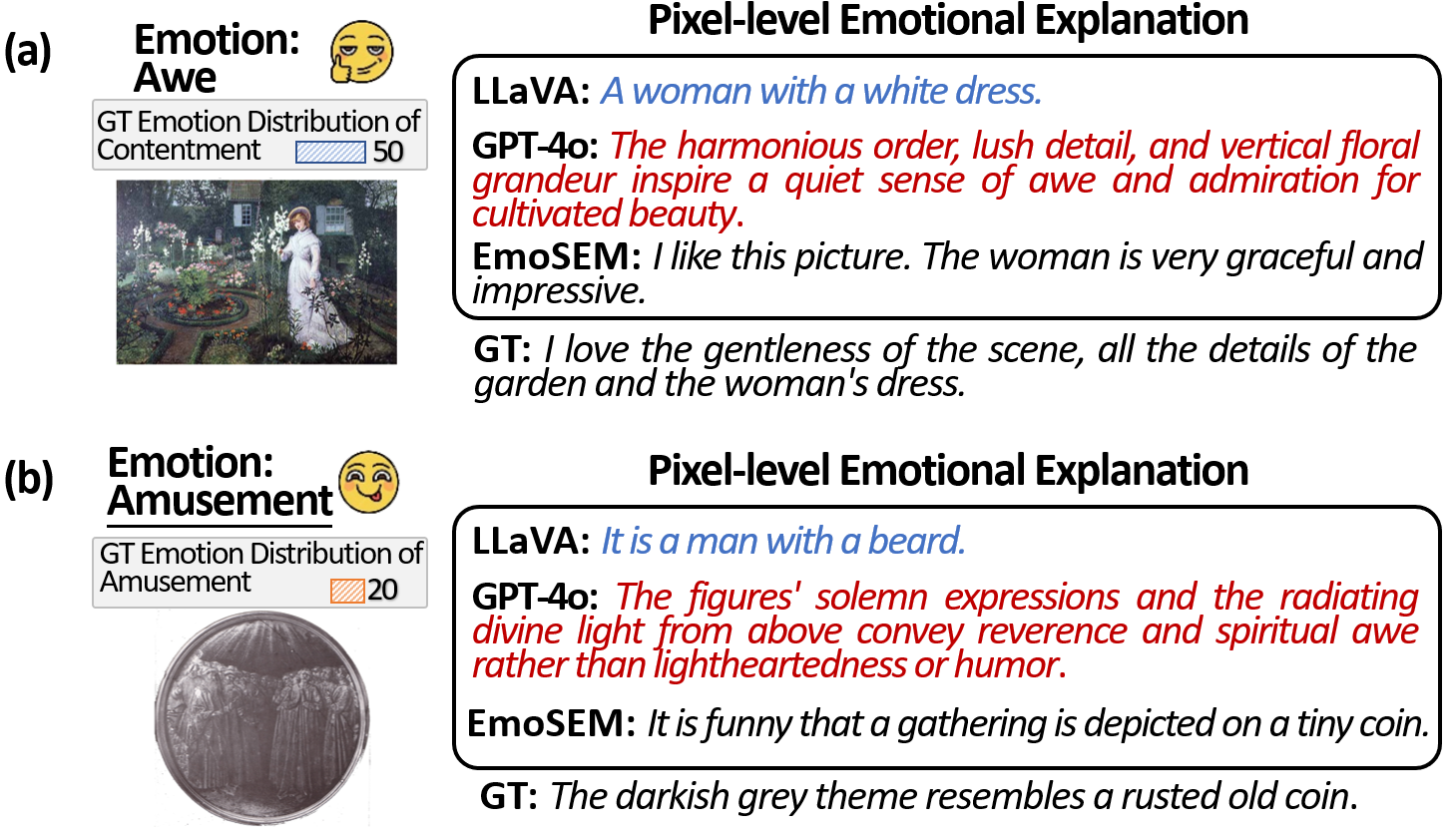}
    \caption{Visualization comparison of pixel-level emotional explanations from our EmoSEM , GPT-4o and LLaVA-7B.}
    \label{fig:qualitative results of mllms}
    \vspace{-0.3mm}
\end{figure}

\textbf{GPT-4o and LLaVA-7B.} To assess their capability for pixel-level emotional explanation, we present MLLMs, including GPT-4o and LLaVA-7B, with the prompt, \ie, ``\textit{Which pixel areas of visual art evoke \_\_? Output a brief explanation of why these pixel areas are evocative}'' . The qualitative results are shown in Fig.~\ref{fig:qualitative results of mllms}. 

Our observations are as follows: 1) LLaVA-7B generally provides explanations that remain at the level of \textbf{objective description}, such as ``a woman with a white dress'' in Fig.~\ref{fig:qualitative results of mllms}~(a) and ``it is a man with a bread'' in Fig.~\ref{fig:qualitative results of mllms}~(b), lacking emotional reasoning. 2) GPT-4o exhibits two major limitations: \textbf{First, its generated explanations tend to be linguistically ornate yet lack substantive content, resembling a literary aggregation of words.} For example, in Fig.~\ref{fig:qualitative results of mllms} (a), the explanation is ``\emph{the harmonious order}, \emph{lush detail}, and vertical floral \emph{grandeur} inspire a quiet sense of awe and \emph{admiration for cultivated beauty}''. \textbf{Second, it struggles to produce accurate substantive visual stimuli localization for emotion explanations, especially for non-dominant emotions.} For instance, in Fig.~\ref{fig:qualitative results of mllms}~(b), where the target emotion is ``amusement'', GPT-4o focuses on inaccurate visual stimuli context and asserts that the emotion is absent, stating ``\emph{the figures'solemn expressions and the radiating divine light} from above convey reverence and spiritual awe rather than lightheartedness or humor.'' In contrast,  out model explains that ``it is funny that a \emph{gathering} is depicted on a tiny coin'', focusing on the whole scene activity in the coin.

\ul{\emph{\textbf{Overall}, these MLLM models still tend to force explanations that fit the given emotion category based on linguistic style, rather than truly grounding them in visual content. Concretely, when they fail to identify suitable visual targets, they often produce statements that appear plausible but lack visual support, or rely on ornate language without grounding. In contrast, our approach achieves better alignment among visual evidence, emotional intent, and linguistic expression.}}

\begin{figure}
  \centering
\includegraphics[width=0.98\linewidth]{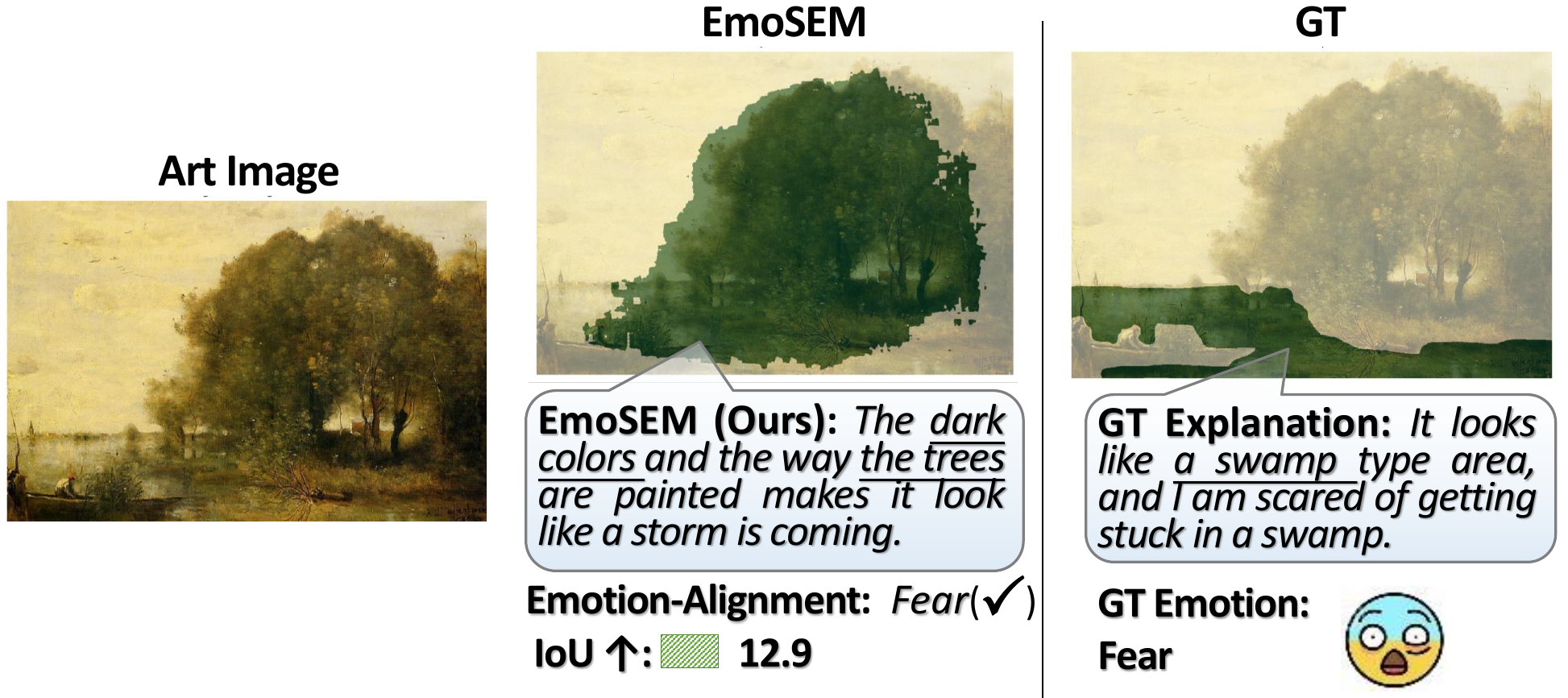}
    \caption{
  Failed case. Our mask differs from that of GT but aligns with the intended emotion and the interpretation remains reasonable.}
    \label{fig:failed-cases}
\end{figure}

\textbf{Failure case.} In Fig.~\ref{fig:failed-cases}, we show a failure sample. The GT accurately localizes ``a swamp'' area with the explanation "It looks like a swamp type area, and I am scared of getting stuck in a swamp". The predicted mask focuses on the color and object ``trees'' with the explanation ``The dark colors and the way the trees are painted makes it look like a storm is coming''. It is worth noting that although there are differences in segmentation accuracy (IoU=12.9) and semantic of interpretation between us and GT, our explanation is semantically consistent with the predicted mask and correctly conveys the emotion of ``Fear''.
This suggests that complex art scenes may contain multiple valid emotional triggers, while the dataset provides only a single ground-truth region. \emph{Such annotation limitations, together with the subjectivity of emotion, can lead to low IoU despite reasonable predictions.} Future work should develop multi-dimensional evaluation frameworks and construct datasets that allow multiple emotional focal points per image, capturing the inherent subjectivity of art and enabling a fairer and more comprehensive assessment of model performance.

\putbib
\end{bibunit}

\end{document}